\documentclass[conference, 9pt]{IEEEtran}

\usepackage{cite}
\usepackage{amsmath,amssymb,amsfonts}
\usepackage{algorithmic}
\usepackage{graphicx}
\usepackage{textcomp}
\usepackage{wrapfig}
\usepackage{lipsum}
\usepackage{graphicx}
\usepackage{subcaption}
\usepackage[font=small,skip=2pt]{caption} 
\usepackage{algorithm}
\usepackage{algorithmic}

\usepackage{threeparttable}
\usepackage{multirow}
\usepackage{multicol}
\usepackage{subcaption} 
\usepackage{amsmath}
\usepackage{circledtext}
\usepackage{tikz}
\usepackage{url}
\usepackage{graphicx}
\usepackage{float}
\usepackage{placeins}
\usepackage{lipsum}
\usepackage{url}
\usepackage[colorlinks]{hyperref}

\hypersetup{linkcolor=blue,
,citecolor=magenta,
,filecolor=Mulberry
}

\usepackage[square,sort,comma,numbers,compress]{natbib}

\usepackage{xcolor}
\def\BibTeX{{\rm B\kern-.05em{\sc i\kern-.025em b}\kern-.08em
    T\kern-.1667em\lower.7ex\hbox{E}\kern-.125emX}}

\begin{document}


\title{ \textbf{AnalogFed}: Privacy-Preserving Discovery of \underline{\textbf{Analog}} Circuits at Scale with \underline{\textbf{Fed}}erated Generative AI}

\author{
    Qiufeng Li, Shu Hong, Tian Lan, Weidong Cao \\
    Department of Electrical and Computer Engineering, The George Washington University, DC, USA 
}




\maketitle

\begin{abstract}
Recent advances in generative AI (GenAI) have shown transformative potential for modern hardware design.
However, existing GenAI-driven approaches fall short of enabling large-scale electronic design automation (EDA) due to the proprietary and siloed nature of hardware datasets, which cannot be centralized for model training.
Achieving at-scale GenAI-driven EDA, therefore, requires a novel privacy-preserving framework that can leverage distributed data without compromising confidentiality.
This work introduces \textbf{AnalogFed}, the first privacy-preserving framework for large-scale analog circuit topology discovery using federated learning (FedL) and GenAI. 
AnalogFed establishes the feasibility of collaborative analog topology design while addressing key security challenges: it mitigates membership inference attacks (MIAs) through a novel input perturbation strategy based on dummy token injection, and defends against model inversion attacks with customized, efficient homomorphic encryption. 
Extensive experiments demonstrate AnalogFed’s effectiveness and efficiency, achieving strong privacy protection without degrading model utility. 
This framework lays the foundation for scalable, multi-party collaboration in next-generation hardware design automation with GenAI.

\end{abstract}

\section{Introduction}

\begin{figure}[!t]
\centering
\includegraphics[width=0.45\textwidth]{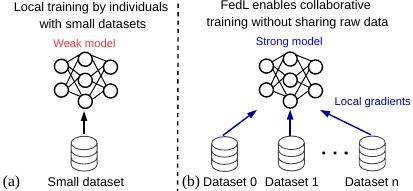}
\caption{Comparison between (a) current centralized training, where individuals train independently on small, isolated datasets, resulting in suboptimal model performance, and (b) Privacy-preserving Federated Learning, which enables collaborative training via model parameter exchange while maintaining strict data locality and confidentiality.  }
\label{FL}
\end{figure}

Generative AI (GenAI)-driven electronic design automation (EDA) is reshaping modern hardware design~\cite{analogcoder, chipnemo, ho2025verilogcoder, gao2025analoggenie}, enabling tasks ranging from generating versatile digital modules to discovering unseen analog circuits.
Despite its great promise, GenAI for hardware design still lags far behind its success in language~\cite{genai} and vision~\cite{cv}. 
One primary bottleneck is the \textbf{data privacy}:
unlike the immense open corpora used to train superb language and vision foundation models, practical hardware datasets (e.g., Verilog codebases and analog circuit libraries) are often proprietary, siloed across organizations, and inherently non-centralizable~\cite{circuitnet}.
Thus, today's GenAI-driven EDA research is often limited to small datasets curated by individual researchers (Fig.~\ref{FL}(a)) and struggles to unleash the full potential of generative innovation that relies on big data~\cite{circuitnet}.
To unlock the next leap in hardware design productivity, performance, and innovation with GenAI-driven EDA, there is an urgent need for a novel, privacy-preserving means to leverage distributed private data for collaborative model training, enabling design automation at scale.

Federated learning (FedL)~\cite{fedavg} enables collaborative training across institutions without sharing raw data (Fig.~\ref{FL}(b)), making it a natural fit for GenAI-driven hardware design, where datasets are typically proprietary and siloed.
Specifically, FedL updates global models by aggregating local gradients across distributed data sources (companies, design houses, or institutions), thereby producing models that are significantly more capable than those trained on any single local dataset.
Yet, despite avoiding direct data sharing, FedL still poses significant privacy and security risks due to the exchange of model parameters.
This communication channel exposes indirect attack surfaces, including model inversion~\cite{inversion} and membership inference attacks (MIAs)~\cite{mia_survey}.
Model inversion can reconstruct proprietary design artifacts from transmitted plain gradients during FedL (Fig.~\ref{overview}(c)).
MIAs can reveal whether specific circuits or design elements were included in the training set after the model is distributed (Fig.~\ref{overview}(d)).
Specifically, GenAI models are particularly vulnerable to MIAs~\cite{sue_openai, oren2023proving, books_shutdown}, because their autoregressive nature encourages memorization and high-confidence next-token prediction.
However, existing defenses are limited and largely inadequate~\cite{mia_survey}.
Consequently, applying FedL to GenAI-driven hardware design faces unique privacy and security vulnerabilities.
These gaps highlight the pressing need to develop domain-specific FedL to scale GenAI-driven EDA securely.

This work introduces the first domain-specific FedL framework for scalable GenAI-driven EDA with strong privacy preservation.
In particular, we focus on the challenging problem of analog circuit topology discovery, where design automation remains limited and public datasets are far scarcer than in digital design, making federated collaboration especially valuable. 
We term this framework \textbf{AnalogFed}, which not only establishes the feasibility of collaborative analog topology design but also comprehensively addresses both the heightened risk of MIAs in circuit generative models and the conventional vulnerabilities in federated training. 
AnalogFed introduces a novel input perturbation mechanism to mitigate MIAs, injecting carefully selected dummy tokens to reduce token-level overfamiliarity while excluding them from the loss function to preserve model utility. 
It further tailors efficient homomorphic encryption (HE)~\cite{fedml, dict} to defend against model inversion attacks.
Together, these strategies make AnalogFed a scalable, secure framework for discovering various analog circuits using GenAI.
Our work paves the way for future at-scale GenAI-driven EDA through multi-party collaborations.

Our key contributions are as follows:


\begin{itemize}

\item \textbf{First FedL Framework for GenAI-Driven EDA.}
We pioneer the integration of FedL with GenAI for analog design automation, laying a foundation for future at-scale, privacy-preserving EDA and collaborative hardware design.

\item \textbf{Two-level Privacy-Preserving Collaboration.} 
To ensure both data confidentiality and deployment flexibility, we propose a two-level privacy framework that protects proprietary topology datasets from analog designers and commercial fabrication technology from manufacturers.

\item \textbf{Domain-Specific FedL Defense Mechanisms.} 
We introduce a novel dummy-token input perturbation strategy to mitigate MIAs with minimal overhead. 
In addition, we optimize HE by encrypting only the initial significant gradient updates during FedL.
These techniques provide strong privacy while maintaining high efficiency and model utility.

\item \textbf{Comprehensive Evaluation on Feasibility, Scalability, and Security.} We conduct extensive experiments across diverse federated settings, demonstrating the feasibility, scalability, and robustness of AnalogFed in analog circuit topology discovery. 
Our evaluation reveals optimized communication and computation overhead, as well as superior defense without hurting modeling utility.

\end{itemize}

\section{Background and Related Work}

\subsection{GenAI for Hardware Design}

GenAI is transforming hardware design across all domains%
~\cite{genai, qwen2.5, openai4o, analogcoder, ho2025verilogcoder}, 
achieving capabilities that surpass traditional methods. 
For example, NVIDIA’s ChipNeMo%
~\cite{chipnemo}, a domain-specialized large language model (LLM), 
can rapidly generate functional digital designs from only a few prompts.
Nonetheless, GenAI's emerging application to analog design automation is even more compelling, as analog design inherently involves abstraction and modeling complexities, thereby heavily relying on time-consuming, expertise-driven manual iterations.
Among all design stages, analog topology design is particularly challenging: it requires conceiving new circuit structures from scratch and largely determines final circuit performance.
GenAI has recently shown great promise in tackling this task, generally falling into two categories: prompt engineering with general-purpose LLMs and domain-specific models.
Early methods such as AnalogCoder~\cite{analogcoder} and AnalogXpert~\cite{analogxpert} have explored prompt engineering for analog circuit generation.
However, general-purpose LLMs are limited in their ability to produce high-quality, diverse analog circuits, as they are not primarily trained for circuit design.
In contrast, domain-specific models, such as CktGNN~\cite{cktgnn} and LaMAGIC~\cite{lamagic} achieve high-quality generation but are limited to specific analog circuits.
Most recently, AnalogGenie~\cite{gao2025analoggenie} has shown remarkable generation performance by broadening the variety of analog circuits, increasing the number of devices within a single design, and discovering previously unseen circuit topologies, thereby surpassing the prior art. 

Despite these advances, existing GenAI approaches for hardware design are fundamentally limited in scalability: they rely on small, centralized datasets and cannot leverage distributed private design data across organizations.
The publicly available datasets of analog circuit design are exceedingly scarce compared to their digital counterparts, making this challenge particularly severe.
This underscores the urgent need for a privacy-preserving and collaborative framework to enable at-scale GenAI-driven analog circuit design.



 \subsection{Privacy-Preserving Machine Learning}
Federated learning (FedL) is a distributed machine learning paradigm that enables multiple clients to collaboratively train a shared model without exposing their local data and instead exchanging model updates (e.g., gradients) with a central server.
It has been widely used in domains such as finance and healthcare, where data is sensitive and decentralized~\cite{fedavg, fedbn, fedcv, fl_healthcare, fl_nlp, scaffold}.
This makes it natural for scaling data-driven analog EDA by harnessing proprietary, distributed circuit datasets.
However, FedL alone does not ensure robust privacy and remains susceptible to critical threats, such as (i) model inversion attacks that reconstruct proprietary design artifacts (e.g., circuits) from transmitted gradients~\cite{inversion}, and (ii) MIAs that aim to reveal whether sensitive training data (e.g., circuits) were part of the training set based on model outputs~\cite{mia_survey}.
GenAI-driven hardware design is especially vulnerable to these security issues under the FedL setting.
Due to their autoregressive nature, GenAI models inherently encourage memorization and high-confidence next-token prediction, making them particularly prone to stronger MIAs. 
Numerous studies have confirmed this vulnerability~\cite{mia_survey,loss, conrecall, recall, mink, mink++, extract_data}, yet effective defenses remain scarce.

Conventional defenses such as pruning and regularization mainly reduce overfitting but offer only limited protection for GenAI models~\cite{regul_defense}.
Differential privacy (DP) offers formal guarantees and has been applied to FedL in many forms~\cite{dp_defense,dldp, lightdpfl, dpgdp}, but the required noise levels typically cause substantial utility degradation for large autoregressive models. 
SOFT~\cite{soft} mitigates MIAs by replacing fine-tuning data with responses from commercial LLMs, yet this strategy is unsuitable for analog circuit design due to domain-specific representation requirements. 
These limitations underscore the need for domain-specific challenges and customized defense mechanisms when applying FedL to GenAI-driven hardware design (i.e., analog EDA in this work).

\section{AnalogFed Framework}





\begin{figure*}[!t]
\centering
\includegraphics[width=0.99\textwidth]{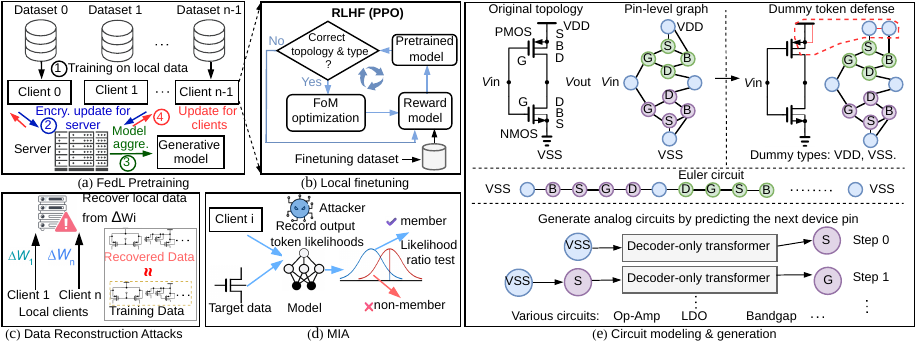}
\caption{\textbf{Overview of the AnalogFed framework.} (a) Federated pre-training for privacy-preserving collaborative learning. (b) Client-side fine-tuning via RLHF (PPO) for design optimization. Threat models: Data reconstruction (model inversion) from shared gradients (c) and MIA targeting training data (d). (e) Circuit generation using a decoder-only Transformer with Euler-tokenized topologies and dummy token injection.}
\label{overview}
\end{figure*}


AnalogFed focuses on collaborative discovery of analog circuit topologies while preserving privacy.
With the end of Moore’s Law~\cite{moore}, further performance scaling of analog circuits can no longer rely on device miniaturization alone. 
Discovering novel analog topologies has thus become a promising pathway for post-Moore performance gains, analogous to discovering new protein structures to unlock improved biological functions.
Ideally, a GenAI approach to designing circuit topologies would use a comprehensive dataset of all analog circuit types to train a powerful, centralized model that can generate arbitrary circuit topologies.
Yet, this ideal scenario is unlikely to be realized in the real world, as analog circuit developers and EDA researchers are unwilling to share their private data to build a large dataset due to concerns over intellectual property (IP).
A practical case is to use FedL to collaboratively train the model without directly accessing distributed private topology datasets from individual clients.
AnalogFed is proposed for this critical need.
In the following subsections, we will present the key challenges of AnalogFed and our proposed techniques to address them.

\subsection{Deployment Challenge, Threat Model, and Framework Overview}

\noindent{\textbf{Deployment Challenge.}} In the FedL setting, analog circuit discovery entails multi-faceted privacy issues.
Clients must not only protect their IP (i.e., circuit topologies) but also safeguard the semiconductor fabrication technologies provided by manufacturers that are used to evaluate the performance of circuit topologies.
This domain-specific layered complexity poses a significant barrier to the practical adoption of FedL, raising a critical question: how can we leverage AnalogFed to jointly protect designer IP and semiconductor technology?

\noindent{\textbf{Threat Models.}} In addition to the deployment challenge, FedL faces threats during learning and after learning processes.
In terms of threat models, we assume a semi-honest server that strictly adheres to the FedAvg protocol~\cite{fedavg} but seeks to reconstruct private circuit topologies from transmitted gradients, as shown in Fig.~\ref{overview}(c). 
Additionally, we consider a black-box malicious client who, upon receiving the global model, attempts to launch MIAs by evaluating the model's prediction confidence on target circuit sequences. The attacker’s goal is to distinguish between member circuits (used in federated training) and non-member circuits, as shown in Fig.~\ref{overview}(d).

\noindent\textbf{{Framework Overview.}} AnalogFed addresses the deployment challenge and threats by integrating three key techniques: (1) a two-stage deployment strategy to protect the privacy of both designers and manufacturers (Section~\ref{sec: deployment}), (2)  secure gradient aggregation (Section~\ref{sec: HE}), and (3) novel input perturbations to defend against MIAs while preserving model utility (Section~\ref{sec: dp}).
The central tenet of AnalogFed is to \textbf{offer strong security and high efficiency without compromising model utility}.

\subsection{Two-Stage Privacy-Preserving Deployment}
\label{sec: deployment}

We propose a two-stage privacy-preserving deployment strategy: federated pre-training and decentralized fine-tuning, as shown in Figs.~\ref{overview}(a) and (b).
Specifically, within AnalogFed, clients collaboratively pre-train a shared generative model, enabling the model to capture foundational knowledge of circuit connectivity across distributed analog topologies.
Each client will then perform fine-tuning on a private dataset for specific tasks, with circuit topologies evaluated locally to protect semiconductor technology.
This strategy is strategically attached to the practice: 
\textit{Individual clients can collaboratively pre-train a powerful model by using knowledge distilled from all clients’ private datasets--significantly outperforming models trained on any single client’s limited data. 
Then, clients can locally fine-tune the model to identify high-performance circuits of interest by utilizing a specific semiconductor fabrication technology, thereby ensuring the privacy of semiconductor processes.}

\subsubsection{\textbf{\textit{FedL Setting for Pre-training}}}

We adopt a horizontal federated learning setting comprising one central server and $n$ clients.
As shown in Fig.~\ref{overview}(a), each client performs a round of local training (with input perturbations proposed in Section~\ref{sec: dp}) on its own dataset (\tikz[baseline=(char.base)]{
    \node[shape=circle, draw, inner sep=0.5pt] (char) {1};
}) and then sends the resulting model updates (encrypted by using the technique in Section~\ref{sec: HE}) to a server (global aggregator, \tikz[baseline=(char.base)]{
    \node[shape=circle, draw, inner sep=0.5pt] (char) {2};
}).
The server combines these updates using an algorithm such as Federated Averaging (\textit{FedAvg} protocol~\cite{fedavg}, \tikz[baseline=(char.base)]{
    \node[shape=circle, draw, inner sep=0.5pt] (char) {3};
}), and redistributes the aggregated model updates to all clients for the next round (\tikz[baseline=(char.base)]{
    \node[shape=circle, draw, inner sep=0.5pt] (char) {4};
}). 
We assume a \textit{semi-honest} (honest-but-curious) server that follows the prescribed training protocol without tampering but attempts to infer membership information from client updates. 

While there are several domain-specific models, we use the \textit{AnalogGenie}~\cite{gao2025analoggenie}, a state-of-the-art open-source GenAI model for analog topology generation, as our backbone. 
LaMAGIC~\cite{lamagic} is from IBM, which is not open source.
At the core of AnalogGenie is a graph representation of analog circuit topologies (top of Fig.~\ref{overview}(e)), which models each device pin (e.g., gate  (G), drain (D), source (S), and bulk (B) of a transistor) as an individual graph node.
This pin-level granularity ensures that all electrical connections and inter-device interactions are explicitly represented, offering a cutting-edge, precise structural encoding universal to all analog circuit topologies. 
A circuit graph is subsequently converted into an Eulerian circuit, which is a sequence of traversal paths of the circuit graph starting at the ``\texttt{VSS}'' token, a default device pin for power supply, used by every analog circuit.
AnalogGenie formulates topology design as sequence generation by treating each device-pin connection as a token and predicting the next pin using a domain-specific autoregressive decoder (bottom of Fig.~\ref{overview}(e)).
This autoregressive training makes it vulnerable to MIAs, as revealed in Section~\ref{sec: dp}.

\subsubsection{\textbf{\textit{Local Fine-Tuning}}}

In practice, a client is often interested in developing a specific type of analog circuit with a particular semiconductor process.
Thus, the client can utilize a set of performance-labeled topologies aligned with their circuit type, specification objectives, and semiconductor process preferences to fine-tune the pre-trained model for local circuit discovery.
We propose a fine-tuning strategy that uses the reinforcement learning with human feedback (RLHF) method based on Proximal Policy Optimization (PPO)~\cite{humanfeedback}, given RLHF's stability and success in fine-tuning LLMs. 
For example, if a client (i.e., designer) aims to develop/discover high-performance operational-amplifiers (Op-Amp), the client can prepare a small annotated dataset (e.g., labeled Op-Amps) with performance metrics like figure-of-merit (FoM) and train a reward model (Fig.~\ref{overview}(b)) as a proxy for human judgment.
An example of training a reward model can be found in prior work~\cite{reward}.
With this rewarding process, PPO then updates the model to prioritize topologies with higher FoMs.
Through this process, the model will converge to generate novel, high-performance analog topologies, pushing the boundaries of conventional human designs.

\subsection{Dynamic Homomorphic Encryption}
\label{sec: HE}

By exploiting gradients associated with specific data samples and model weights, an attacker can potentially reconstruct the original input data. 
In a conventional FedL setting, each local client iteratively computes and uploads plain-text gradients to the server.
The central server aggregates local gradients over training iterations, enabling it to infer complete model weights and thus launch inversion attacks.
Homomorphic Encryption (HE)~\cite{openfhe} has emerged as the most effective means, enabling secure aggregation directly on encrypted data.
In this approach, clients encrypt model gradients into ciphertexts, and the additive property of HE guarantees that the encryption of a sum equals the product of individual ciphertexts (e.g., $Enc(w_1)\times Enc(w_2)=Enc(w_1+w_2)$). 
Yet, conventional HE approaches introduce substantial computational and communication overhead because all model updates are encrypted throughout the entire FedL process.
Recent work has shown that encrypting only a subset of updates~\cite{dict} in each epoch can effectively mitigate model inversion attacks.
Nevertheless, this selective encryption strategy is limited to FedL fine-tuning with low-rank adaptation for specific tasks and thus is not suitable for FedL-based pre-training in general analog circuit discovery.

We propose a {Dynamic Homomorphic Encryption (DHE)} mechanism that adaptively encrypts gradients during FedL pre-training. 
Our method is inspired by the well-observed convergence behavior in deep neural networks, where the magnitudes of gradients naturally decay as training progresses.
During the early training stages, gradients are large due to high loss values and significant model updates, making them more informative to the final weights.
In contrast, during the latter training, gradient magnitudes become much smaller and contribute less to the final weights (Fig.~\ref{commun}(a)). 
Leveraging this observation, DHE selectively encrypts the early, large-magnitude gradients to effectively prevent adversaries from reconstructing model parameters, thereby mitigating model inversion attacks. 
To balance security and efficiency, DHE dynamically adjusts encryption based on convergence: when the loss reduction remains below 20\% for five consecutive epochs, encryption is automatically disabled. 
In our experiments, encryption is typically required for the first 600 rounds (Fig.~\ref{commun}(b)). 
Unlike prior approaches that statically encrypt a fixed percentage of sensitive gradients (e.g., FEDML-HE~\cite{fedml}), DHE adaptively focuses on the most informative and vulnerable training stages, substantially reducing both computational and communication overhead while preserving privacy.

\subsection{Few Dummy Token, Strong Privacy}
\label{sec: dp}

\begin{table}[!t]
\centering
\caption{Effectiveness of traditional defense against multiple MIAs. Performance is measured using AUC-ROC scores, where lower values indicate stronger defense.}
\setlength{\tabcolsep}{1mm}
\begin{tabular}{cccccc}
\hline
MIAs       & Pretrain & Pruning & Regul. & GRIP  & DP    \\ \hline
Loss       & 0.839    & 0.735   & 0.826  & 0.745 & 0.521 \\
Zlib       & 0.814    & 0.716   & 0.810  & 0.734 & 0.516 \\
Mink       & 0.839    & 0.728   & 0.821  & 0.724 & 0.538 \\
Mink++     & 0.793    & 0.724   & 0.786  & 0.729 & 0.549 \\
Ratio      & 0.816    & 0.725   & 0.812  & 0.738 & 0.543 \\
ReCall     & 0.807    & 0.733   & 0.814  & 0.737 & 0.535 \\
CON-ReCall & 0.825    & 0.728   & 0.821  & 0.744 & 0.55  \\ \hline
Test Loss  & 0.094    & 0.103   & 0.101  & 0.104 & 2.45  \\ \hline
\end{tabular}
\label{tab:trans_defense}
\end{table}

\begin{algorithm}[t]
\caption{Pretraining with dummy token insertion.}
\label{alg:analogfed_dummy}
\small
\begin{algorithmic}[1]
\REQUIRE Global model $M_w$, dummy token set $V_D$, dummy ratio $\rho$, client set $\mathcal{C}$. Each client $c \in \mathcal{C}$ has a local analog circuit dataset $\mathcal{D}_c$.
\ENSURE Privacy-preserved global model $M_w$.

\STATE Server initializes global weights $w_0$.
\FOR{each round $t = 1, \dots, T$}
    \STATE Server broadcasts current weights $w_t$ to all clients $c \in \mathcal{C}$.
    \FOR{each client $c \in \mathcal{C}$ \textbf{in parallel}}
        \STATE Initialize perturbed circuit dataset $\mathcal{D}'_c \gets \emptyset$.
        \FOR{each circuit sequence $s_i \in \mathcal{D}_c$}
            \STATE Calculate number of dummy tokens $n = \lfloor |s_i| \cdot \rho \rfloor$.
            \STATE Randomly select $n$ positions within the sequence $s_i$.
            \STATE Inject $n$ tokens from $V_D$ into $s_i$ to generate perturbed sequence $s'_i$.
            \STATE $\mathcal{D}'_c \gets \mathcal{D}'_c \cup \{s'_i\}$.
        \ENDFOR
        \STATE Train local model $M_{w_c}$ on $\mathcal{D}'_c$ for $E$ epochs starting from $w_t$.
        \STATE Upload local weights $w_{c,t}$ to the server.
    \ENDFOR
    \STATE Server performs aggregation: $w_{t+1} \gets \sum_{c \in \mathcal{C}} \frac{|\mathcal{D}_c|}{|\mathcal{D}|} w_{c,t}$.
\ENDFOR
\end{algorithmic}
\end{algorithm}

The proposed dynamic HE mechanism safeguards model privacy on the server side during FedL.
Yet, once the global model is distributed, a malicious client may still launch MIAs by analyzing the model’s output behavior (e.g., confidence scores or logits) before and after aggregation.
Autoregressive GenAI models, such as AnalogGenie~\cite{gao2025analoggenie}, are particularly susceptible to MIAs because their iterative training process often leads to the memorization of training patterns.
Recent studies show that most emerging MIAs exploit this behavior, inferring client-sensitive data from model confidence, where higher confidence typically signals training membership~\cite{extract_data}.
We evaluate several MIA strategies on the AnalogGenie backbone using Area Under the
Receiver Operating Characteristic Curve (AUC-ROC), which measures the model’s discriminative power between member and non-member samples.
The results, summarized in Table~\ref{tab:trans_defense}, reveal that traditional defenses, including pruning, regularization, and GRIP~\cite{regul_defense}, provide limited protection in large-scale generative contexts.
While differential privacy (DP) offers stronger resistance by injecting noise into model updates, it often leads to substantial utility degradation (i.e., large loss).
These observations highlight the need for domain-adaptive privacy defenses.

To defend MIAs, we propose shifting the training data distribution to reduce the model’s familiarity with targeted attack inputs.
In analog circuit topologies, power supply nodes such as \texttt{VDD} (power) and \texttt{VSS} (ground) are ubiquitous, with every transistor typically connected to at least one of them, for example, the bulk pin of a PMOS connected to \texttt{VDD}, and that of an NMOS connected to \texttt{VSS}.
Leveraging this structural property, we introduce dummy wires that connect specific nodes to \texttt{VDD} or \texttt{VSS} (see an example in the top of Fig.~\ref{overview}(c)) to perturb the original topology without altering its functionality.
These dummy connections mirror industry-standard practices of inserting dummy components for layout symmetry and device matching, but are repurposed here as a lightweight, domain-specific privacy defense.
We adopt this dummy connection strategy as our primary defense mechanism.
Note that this strategy differs from conventional DP, which introduces noise into model updates; instead, we add noise (dummy tokens) to the input.

We inject dummy-token blocks after occurrences of \texttt{VSS} and \texttt{VDD} at a specified ratio, perturbing inputs without altering the original schematics. During training, we use the standard autoregressive loss but mask targets that are dummy tokens. Let $\mathbf{X}_{1:N}$ be the original sequence and $\tilde{\mathbf{X}}_{1:N+M_d}$ the sequence with $M_d$ inserted dummy tokens. Define $S_{\text{ignore}}=\{k:\tilde{x}_{k+1}=\texttt{<dummy>}\}$. For a decoder-only Transformer $f$ (e.g., AnalogGenie), the modified loss is
\begin{equation}
\mathcal{L}(f,\tilde{\mathbf{X}}_{1:N+M_d})
=\sum_{\substack{k=1\\ k\notin S_{\text{ignore}}}}^{N+M_d-1}
\mathcal{L}_{\text{CE}}\!\big(\tilde{x}_{k+1},\,f(\tilde{\mathbf{X}}_{1:k})\big).
\label{eq:dummy_loss}
\end{equation}
Thus, dummy tokens act only as input perturbations and do not contribute gradients. This preserves generation quality while improving robustness to MIAs.
The algorithm is shown in Algorithm~\ref{alg:analogfed_dummy}.

 \section{Experiments and Results}

\begin{table*}[t]
\centering
\caption{Embedding Mean and MSE of each client across 3 rounds.}
\begin{tabular}{cllllll}
\hline
\multirow{2}{*}{Round} & \multicolumn{6}{c}{Client (Mean / MSE)}                                                                                                       \\ \cline{2-7} 
                       & \multicolumn{1}{c}{0} & \multicolumn{1}{c}{1} & \multicolumn{1}{c}{2} & \multicolumn{1}{c}{3} & \multicolumn{1}{c}{4} & \multicolumn{1}{c}{5} \\ \hline
1                      & 0.00144 / 0.00040   & 0.00144 / 0.00040   & 0.0014 / 0.0004   & 0.00143 / 0.00040   & 0.00142 / 0.00040   & 0.00143 / 0.00040   \\ \hline
2                      & 0.00145 / 0.00040   & 0.00142 / 0.00040   & 0.00143 / 0.00040   & 0.00143 / 0.00040   & 0.00143 / 0.00040   & 0.00144 / 0.00040   \\ \hline
3                      & 0.00144 / 0.00040   & 0.00143 / 0.00040   & 0.00144 / 0.00040   & 0.00144 / 0.00040   & 0.00143 / 0.00040   & 0.00143 / 0.00040  \\ \hline
\end{tabular}
\label{tab:embedding_stats}
\end{table*}

\begin{table*}[t]
\centering
\caption{Discovery quality comparison across different datasets and client settings.}
\label{tab:evaluation_comparison}

\begin{tabular}{lccccccc}
\hline
\multicolumn{1}{c|}{\multirow{2}{*}{Evaluation Metric}} & \multirow{2}{*}{Validity(\%)} & \multirow{2}{*}{Scalability} & \multirow{2}{*}{Versatility} & \multicolumn{2}{c}{Novelty} & \multicolumn{2}{c}{FoM}   \\ \cline{5-8} 
\multicolumn{1}{c|}{}                                   &                               &                              &                              & Diff Circuit(\%)  & MMD     & Op-Amp  & Power Converter \\ \hline
Centralized (no dummy tokens)                                            & 95.5                          & 63                           & 11                           & 98.6              & 0.0406  & 13744.7 & 3.30            \\
3-Client                                                & 55.7                          & 63                           & 11                           & 90.1              & 0.0768  & 7716.4  & 2.02            \\
6-Client                                                & 65.0                          & 63                           & 11                           & 94.2              & 0.0715  & 10027.4 & 2.35            \\
9-Client                                                & 80.7                          & 63                           & 11                           & 93.2              & 0.0540  & 10007.9 & 2.56            \\
12-Client                                               & 85.6                          & 63                           & 11                           & 96.9              & 0.0533  & 10100.2 & 3.01            \\
16-Client                                               & 91.2                          & 63                           & 11                           & 94.4              & 0.0547  & 11211.3 & 3.10            \\
12-Client-Un                                             & 85.4                          & 63                           & 11                           & 96.6              & 0.0536  & 10206.8 & 2.75            \\
16-Client-Un                                             & 92.4                          & 63                           & 11                           & 95.5              & 0.0508  & 11105.4 & 3.05            \\ \hline
\end{tabular}
\label{tab: fed_discovery}
\end{table*}

\subsection{Experiment Set-up}

\noindent{\textbf{Dataset, Model, and Training.}} We use the dataset of AnalogGenie~\cite{gao2025analoggenie} to evaluate our framework. The dataset comprises 3,350 unique analog circuit topologies spanning 11 functional types: operational-amplifiers (Op-Amps), low dropout regulators (LDOs), bandgap references, comparators, power amplifiers (PAs), power converters, etc, the largest open-source topology datasets to date. 
We use the open-source AnalogGenie model~\cite{gao2025analoggenie} as the backbone for our study, a decoder-only transformer with 6 layers, 6 attention heads, a vocabulary size of 1,029, and approximately 11.8M parameters.
It is the state-of-the-art model for discovering analog circuit topologies.
During pre-training, the dataset is split into 90\% for training and 10\% for validation. 
Each topology is augmented by generating multiple Eulerian paths using graph traversal strategies to improve data diversity, as suggested by AnalogGenie~\cite{gao2025analoggenie}.


\noindent{\textbf{Evaluation Processes and Metrics.}}
We focus on evaluating FedL performance, DHE efficacy and efficiency, and the MIA defense with dummy tokens (Section~\ref{sec: fedl}).
For each evaluation, we provide some baselines or metrics.
\textbf{(A) FedL evaluation:} We assess the generation quality/discovery capability of the model under different FedL settings based on the metrics used by AnalogGenie.
\textbf{(i) Validity:} The percentage of generated topologies that pass SPICE-level syntax and connectivity checks.
\textbf{(ii) Novelty:} The percentage of unique generated circuits that differ from the training data. We quantify topology differences via maximum mean discrepancy (MMD)~\cite{mmd} between generated and real-world circuit graphs.
\textbf{(iii) Performance:} The figure-of-merit (FoM) that considers the key circuit specifications.
\textbf{(iv) Scalability:} The maximum number of devices in the generated topology.
\textbf{(v) Versatility:} The number of distinct analog circuit types it generates.
We compare the performance of FedL to centralized training using pooled datasets. 
\textbf{(B) DHE evaluation:} We assess the efficacy and efficiency of DHE by comparing its communication overhead, training time, and model training loss against the baseline static HE~\cite{fedml}. 
We adopt the CKKS HE scheme with bootstrapping, implemented via OpenFHE~\cite{openfhe}.
We use the same crypto parameters as~\cite{fedml}.
\textbf{(C) MIA defense with dummy tokens:} We evaluate MIA defense using seven standard attacks: \textit{Loss}~\cite{loss}, \textit{Zlib}~\cite{extract_data}, \textit{Mink}~\cite{mink}, \textit{Mink++}~\cite{mink++}, \textit{Ratio}~\cite{extract_data}, \textit{ReCall}~\cite{recall}, and \textit{Con-ReCall}~\cite{conrecall}.
We report MIA success rate with/without our proposed dummy token defense mechanism as a privacy leakage metric, and the Area Under the Receiver Operating Characteristic Curve (AUC-ROC) as the metric. AUC-ROC reflects how well an attacker distinguishes member from non-member data. 
A score of 0.5 implies random guessing, i.e., the best defense. 
In our experiments, we follow prior work~\cite{mink} and use 1,000 samples each for member and non-member data.


\noindent{\textbf{FedL Setup and Training Platform.}} 
For all FedL experiments, we adopt the standard \textit{FedAvg}~\cite{fedavg} algorithm as the aggregation strategy. Each client performs $T = 20$ local training steps per communication round, with a mini-batch size of $B = 64$, over a total of $3500$ rounds. To evaluate scalability, we experiment with 3, 6, 9, 12, and 16 clients. All these FedL experiments are conducted by using our proposed input perturbation mechanism.
The AnalogGenie dataset is partitioned into both balanced and unbalanced settings to reflect realistic deployment conditions. In both scenarios, the data volume scales proportionally with the number of clients. In the unbalanced configuration, each client receives at least $1/32$ of the total dataset. To simulate real-world heterogeneity, we introduce intentional domain specialization among clients. For example, Client 0 primarily focuses on operational amplifiers, Client 1 on bandgap references, and Client 2 on power converters. 
This setup enables a systematic assessment of federated learning robustness under varying data scales and non-IID functional diversities. We evaluate training metrics and circuit generation tasks for each configuration, applying the proposed DHE and dummy token mechanisms to all FedL experiments.
All experiments are performed on a workstation with 8 NVIDIA A6000 GPUs and an AMD EPYC 7313 CPU.

\begin{table*}[t]
\centering
\caption{The effectiveness of our dummy token defense against multiple MIAs. Performance is measured using AUC-ROC scores, where lower values indicate stronger defense.}

\begin{tabular}{clllllllllllllll}
\hline
\multirow{2}{*}{MIAs} & \multicolumn{3}{c}{1\%}                                                  & \multicolumn{3}{c}{2\%}                                                  & \multicolumn{3}{c}{4\%}                                                  & \multicolumn{3}{c}{6\%}                                                  & \multicolumn{3}{c}{8\%}                                                  \\ \cline{2-16} 
                      & \multicolumn{1}{c}{B2} & \multicolumn{1}{c}{B4} & \multicolumn{1}{c}{B6} & \multicolumn{1}{c}{B2} & \multicolumn{1}{c}{B4} & \multicolumn{1}{c}{B6} & \multicolumn{1}{c}{B2} & \multicolumn{1}{c}{B4} & \multicolumn{1}{c}{B6} & \multicolumn{1}{c}{B2} & \multicolumn{1}{c}{B4} & \multicolumn{1}{c}{B6} & \multicolumn{1}{c}{B2} & \multicolumn{1}{c}{B4} & \multicolumn{1}{c}{B6} \\ \hline
Loss                  & 0.769                  & 0.746                  & 0.719                  & 0.655                  & 0.639                  & 0.651                  & 0.592                  & 0.558                  & 0.587                  & \textbf{0.513}         & 0.517                  & 0.549                  & 0.525                  & 0.529                  & 0.574                  \\
Zlib                  & 0.748                  & 0.743                  & 0.723                  & 0.659                  & 0.646                  & 0.650                  & 0.586                  & 0.568                  & 0.555                  & 0.486                  & 0.488                  & 0.485                  & 0.513                  & \textbf{0.499}         & 0.498                  \\
Mink                  & 0.726                  & 0.725                  & 0.723                  & 0.648                  & 0.640                  & 0.642                  & 0.591                  & 0.585                  & 0.593                  & 0.561                  & \textbf{0.541}         & 0.568                  & 0.579                  & 0.572                  & 0.593                  \\
Mink++                & 0.767                  & 0.771                  & 0.720                  & 0.679                  & 0.684                  & 0.671                  & 0.617                  & 0.596                  & 0.612                  & 0.557                  & 0.566                  & \textbf{0.552}         & 0.575                  & 0.559                  & 0.564                  \\
Ratio                 & 0.777                  & 0.772                  & 0.727                  & 0.715                  & 0.767                  & 0.787                  & 0.642                  & 0.614                  & 0.598                  & \textbf{0.542}                  & 0.564                  & 0.558                  & 0.588                  & 0.580                  & 0.618                  \\
ReCall                & 0.769                  & 0.745                  & 0.735                  & 0.697                  & 0.684                  & 0.698                  & 0.627                  & 0.611                  & 0.619                  & 0.557                  & 0.531                  & 0.529                  & 0.523                  & \textbf{0.521}         & 0.528                  \\
CON-ReCall            & 0.776                  & 0.761                  & 0.748                  & 0.712                  & 0.697                  & 0.722                  & 0.652                  & 0.618                  & 0.631                  & 0.552                  & 0.544                  & 0.541                  & \textbf{0.532}                  & 0.534                  & 0.542                  \\ \hline
\end{tabular}
\label{tab:mias}
\end{table*}

\begin{table}[t]
\centering
\caption{The effectiveness of our dummy token defense against multiple MIAs is conducted by performing attacks using dummy-token-injected data as targeted inputs. }
\label{tab:attacks}
\scalebox{0.88}{
\begin{tabular}{ccccccc}
\hline
\multirow{2}{*}{MIAs} & \multicolumn{3}{c}{6\%+B2} & \multicolumn{3}{c}{6\%+B4} \\ \cline{2-7} 
                      & 1\%+B2  & 4\%+B2  & 6\%+B2 & 1\%+B4  & 4\%+B4  & 6\%+B4 \\ \hline
Loss                  & 0.488   & 0.479   & 0.481  & 0.485   & 0.482   & 0.443  \\
Zlib                  & 0.496   & 0.492   & 0.490  & 0.458   & 0.451   & 0.444  \\
Mink                  & 0.474   & 0.481   & 0.475  & 0.484   & 0.463   & 0.465  \\
Mink++                & 0.467   & 0.445   & 0.463  & 0.447   & 0.463   & 0.466  \\
Ratio                 & 0.486   & 0.487   & 0.493  & 0.453   & 0.464   & 0.446  \\
ReCall                & 0.467   & 0.468   & 0.469  & 0.442   & 0.453   & 0.474  \\
CON-ReCall            & 0.475   & 0.474   & 0.471  & 0.464   & 0.469   & 0.468  \\ \hline
\end{tabular} }
\vskip -12pt
\end{table}


 \subsection{Experimental Results}
\label{sec: fedl}

\noindent{\textbf{(A) Characterization of topology data heterogeneity.}} 
Data heterogeneity can hinder FedL, and circuit topology warrants a tailored look. We tokenize circuit graphs and, with a pre-trained encoder, compute per-client embedding \textbf{mean} and \textbf{MSE}. In a six-client setup, Table~\ref{tab:embedding_stats} reports these statistics for one structured (type-based) split and two random partitions; values remain consistent across clients and rounds, indicating stable embeddings. 
We attribute this to the structural regularity of analog circuits, i.e., recurring PMOS/NMOS motifs (current mirrors, differential pairs, gain stages) yield balanced device-type ratios.
So, token compositions stay similar across datasets. 
This low-variance behavior suggests analog circuit datasets are well-suited for FedL, enabling effective collaborative training of robust, generalizable generative models.

\noindent{\textbf{(B) Federated discovery capability.}} To assess the performance and effectiveness of FedL, we evaluate the model’s generation quality and topology discovery capability under varying client configurations, specifically with \textbf{3}, \textbf{6}, \textbf{9}, \textbf{12} and \textbf{16} clients. 
As shown in Table~\ref{tab:evaluation_comparison}, performance consistently improves with an increasing number of clients, highlighting a key advantage of FedL: the ability to train more powerful and generalizable models by leveraging distributed data. 
Importantly, the results show that the proposed dummy token injection strategy does not degrade model utility (compared to the centralized training without dummy tokens), corroborating our hypothesis and aligning with prior findings that structurally inert tokens can support learning~\cite{pause_token}. 
Notably, the unbalanced settings (12-Client-Un and 16-Client-Un), where clients hold datasets of varying sizes, achieve performance comparable to their balanced counterparts.
This demonstrates that FedL remains robust even under non-uniform data distributions, effectively maintaining model quality and stability across heterogeneous client scenarios.


Our experiments demonstrate the feasibility, scalability, and robustness of AnalogFed in analog circuit topology discovery and its promise in addressing collaborative hardware design with distributed private datasets.
This allows clients with limited data to benefit from robust model training without compromising data confidentiality.

\begin{figure}[!t]
\centering
\includegraphics[width=1.0\linewidth]{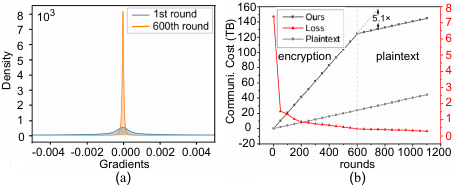}
\caption{(a) Change in gradients distribution, (b) Communication cost (ours vs. plaintext) and training loss over rounds. }
\vskip -12pt
\label{commun}
\end{figure}

\noindent \textbf{(C) DHE Efficiency and Efficacy.}
We evaluate the efficiency of the proposed DHE mechanism using a FedL setup with three clients as an example. 
Following prior work~\cite{fedml}, we apply encryption only during the first 600 rounds and report results for the first 1,100 rounds (approximately 30\% of total training) for comparison.
By encrypting only the early, large-magnitude gradients, DHE effectively reduces communication overhead while maintaining model accuracy.
As shown in Figure~\ref{commun}(b), this selective encryption achieves a \textbf{5.1$\times$ reduction in communication cost}. 
Specifically, a plaintext transmission per round requires 2.03~GB, compared to 10.36~GB for fully encrypted communication in FEDML-HE~\cite{fedml}. 
In terms of training efficiency, plaintext training completes in 8~h, FEDML-HE with 30\% encryption requires 184~h, while our DHE method achieves comparable performance in only 48.5~h.
Yet, further experiments show that DHE provides robust defense capabilities, comparable to FEDML-HE~\cite{fedml} that applies static gradient encryption in FedL.

\noindent \textbf{(D) Dummy Token Defending Capabilities.}
To study the effect of dummy token augmentation, we insert dummy tokens into the training data at ratios of 1\%, 2\%, 4\%, 6\%, and 8\%, using block sizes of 2, 4, and 6 (i.e., consecutive dummy tokens). 
For sequences shorter than 100 tokens, at least one dummy token is guaranteed. 
As shown in Table~\ref{tab:mias}, increasing the dummy ratio improves defense, particularly up to 6\%. 
Beyond the block size of 6, there is little benefit over size 4.
A configuration of 6\% dummy tokens with block size 4 offers the best trade-off between effectiveness and overhead.

We then evaluate robustness when member inputs also contain dummy tokens. 
Using models trained with 6\% dummy tokens (block sizes 2 and 4), we test on inputs with varied dummy configurations. 
As shown in Table~\ref{tab:attacks}, even small mismatches in percentage, block size, or position can lower AUC-ROC, making members indistinguishable from non-members. 
This aligns with prior findings~\cite{small_change} that small input shifts can destabilize predictions.
Overall, dummy token defense renders various MIAs infeasible, making it a novel strategy for privacy-preserving analog circuit design with GenAI.



\subsection{Discussions}
AnalogFed can generalize to various EDA domains constrained by data scarcity and IP protection. Beyond analog topologies, it can be extended to enable privacy-preserving training on proprietary digital RTL codebases~\cite{fed-llm} and process-sensitive radio-frequency (RF) circuit designs~\cite{federated_microwave}.
By utilizing FedL to unlock distributed, siloed datasets, this framework mitigates the ``small data'' challenge in hardware GenAI.
Such a paradigm facilitates the development of foundation-level design agents capable of broad discovery while supporting task-specific fine-tuning for specialized industrial problems.

\section{Conclusion}

This work introduces {AnalogFed}, the first privacy-preserving federated learning framework for large-scale analog circuit topology discovery. Beyond demonstrating the feasibility of collaborative analog generation, AnalogFed systematically addresses key privacy risks inherent in GenAI-driven EDA, including model inversion and MIAs. 
We propose a lightweight dummy-token perturbation mechanism that safeguards design privacy without degrading generative quality. 
Extensive experiments validate its effectiveness, scalability, and efficiency, establishing AnalogFed as a secure and practical foundation for collaborative analog design automation. 
We envision this work as a step toward a broader paradigm of privacy-preserving, cross-organizational GenAI for hardware design at scale.





\vspace{12pt}

\end{document}